\pdfoutput=1
\documentclass[11pt]{article}

\usepackage[]{emnlp2021}

\usepackage{times}
\usepackage{latexsym}
\usepackage{graphics}
\usepackage{graphicx}
\usepackage{multirow}
\usepackage{amssymb}
\usepackage{mathrsfs}%用于产生一种数学用的花体字  
\usepackage{amsmath}%数学符号  
\usepackage{bm}%专门处理数学粗体的bm宏包
\usepackage[cmintegrals]{newtxmath}%高质量数学字体/黑体  
\usepackage{subfigure}
\usepackage{caption3}
\usepackage{makecell}

\usepackage[T1]{fontenc}
\usepackage[utf8]{inputenc}

\usepackage{microtype}

\usepackage{hyperref}

\title{Bridge to Target Domain by Prototypical Contrastive Learning and Label Confusion: Re-explore Zero-Shot Learning for Slot Filling}

\author{Liwen Wang$^{1}$\thanks{\ \ The first two authors contribute equally. Weiran Xu is the corresponding author.} ,
Xuefeng Li$^{1*}$,
Jiachi Liu$^{1}$,
Keqing He$^{2}$,
Yuanmeng Yan$^{1}$,
Weiran Xu$^{1}$\\ 
$^1$Beijing University of Posts and Telecommunications, Beijing, China\\
$^{2}$Meituan Group, Beijing, China\\
\texttt{\{w\_liwen,lixuefeng,ljc1997,yanyuanmeng,xuweiran\}@bupt.edu.cn}\\
\texttt{\{hekeqing\}@meituan.com}
}

\begin{document}
\maketitle
\begin{abstract}
Zero-shot cross-domain slot filling alleviates the data dependence in the case of data scarcity in the target domain, which has aroused extensive research. However, as most of the existing methods do not achieve effective knowledge transfer to the target domain, they just fit the distribution of the seen slot and show poor performance on unseen slot in the target domain. To solve this, we propose a novel approach based on prototypical contrastive learning with a dynamic label confusion strategy for zero-shot slot filling. The prototypical contrastive learning aims to reconstruct the semantic constraints of labels, and we introduce the label confusion strategy to establish the label dependence between the source domains and the target domain on-the-fly. Experimental results show that our model achieves significant improvement on the unseen slots, while also set new state-of-the-arts on slot filling task.\footnote{Our source code is available at: \url{https://github.com/W-lw/PCLC}}
\end{abstract}

\section{Introduction}

Slot filling, as an important part of the task-oriented dialogue system, is mainly used to extract specific information in user utterances. Traditional supervised methods have shown remarkable performance in slot filling tasks \cite{liu2016attentionbased,goo-etal-2018-slot,e-etal-2019-novel,he-etal-2020-learning-tag, wu2020slotrefine,He2020LearningLO,oguz-vu-2020-two,qin2020agif}, but they require a large amount of domain-specific labeled data.

Recently, there has been increasing interest in the zero-shot cross-domain slot filling task, whose goal is to identify unseen slots in the target domain without any labeled data. Previous methods can be classified into two types: one-stage and two-stage. In the one-stage framework, \cite{bapna2017zeroshot,shah2019robust, Lee_Jha_2019} perform slot filling task for each slot type respectively, and the slot type description is then integrated into the prediction process to achieve zero-shot adaptation. The main drawback is that the model possibly predicts multiple slot types for one entity span. To avoid the above problem, \cite{liu2020coach,liu2020zeroresource, he-etal-2020-contrastive} decompose the slot filling task into two stages. First, all slot entities in the utterances are identified by the coarse-grained binary sequence labeling model. Then slot types are classified by mapping the entity value to the representation of the corresponding slot label in the semantic space.

\begin{figure}[t]
\centering
\subfigure[Performance of some zero-shot cross-domain slot filling models on seen and unseen slots in \emph{GetWeather} domain in SNIPS.]{
\resizebox{.44\textwidth}{!}{\includegraphics{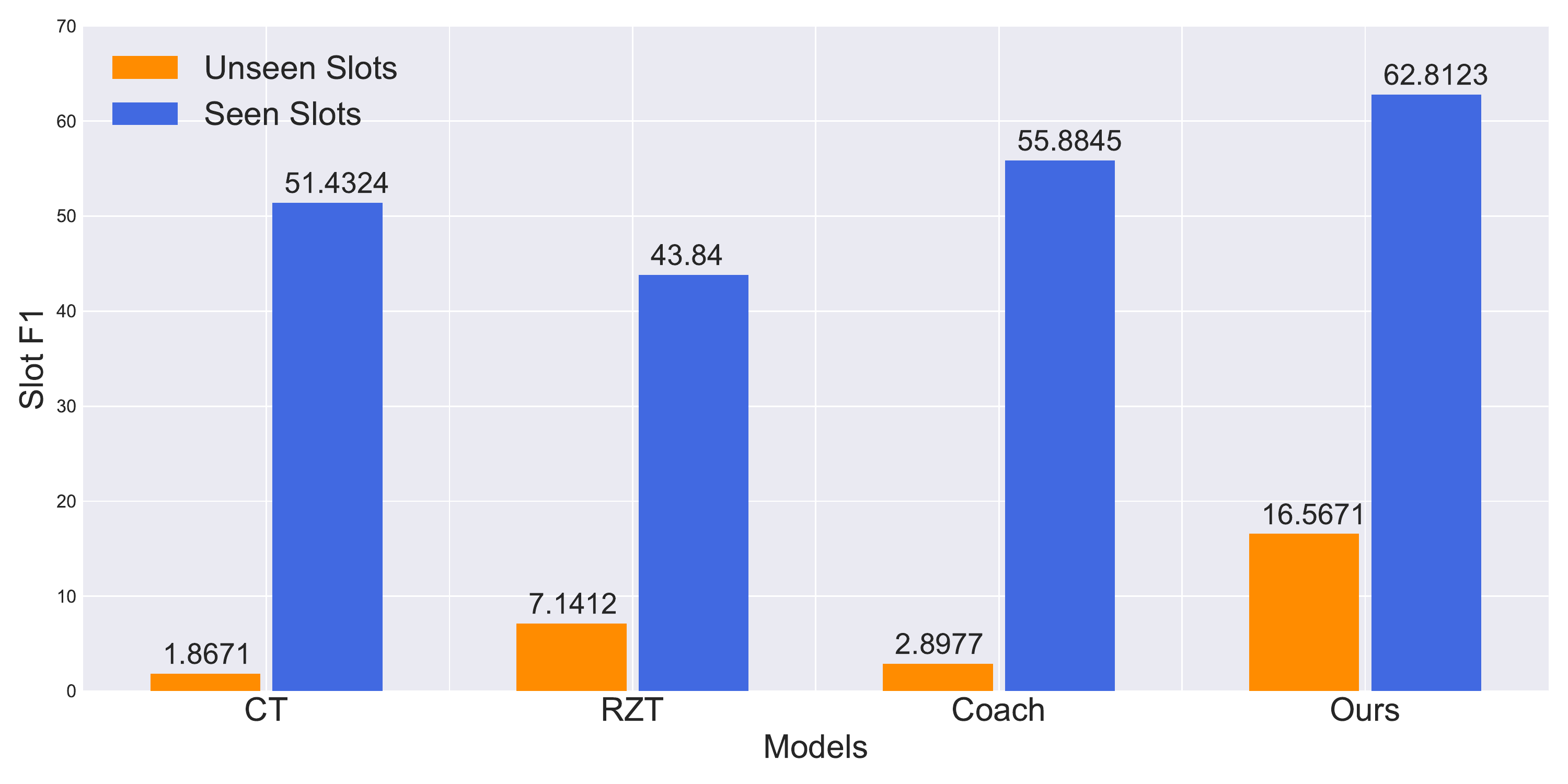}}
}
\subfigure[The left part is the original distribution, where the slot prototypes of the source domain and the target domain are chaotic and lack constraint relations. The right part is the distribution after refinement, where slot prototypes of the source domain and target domain are separated.]{
\resizebox{.46\textwidth}{!}{\includegraphics{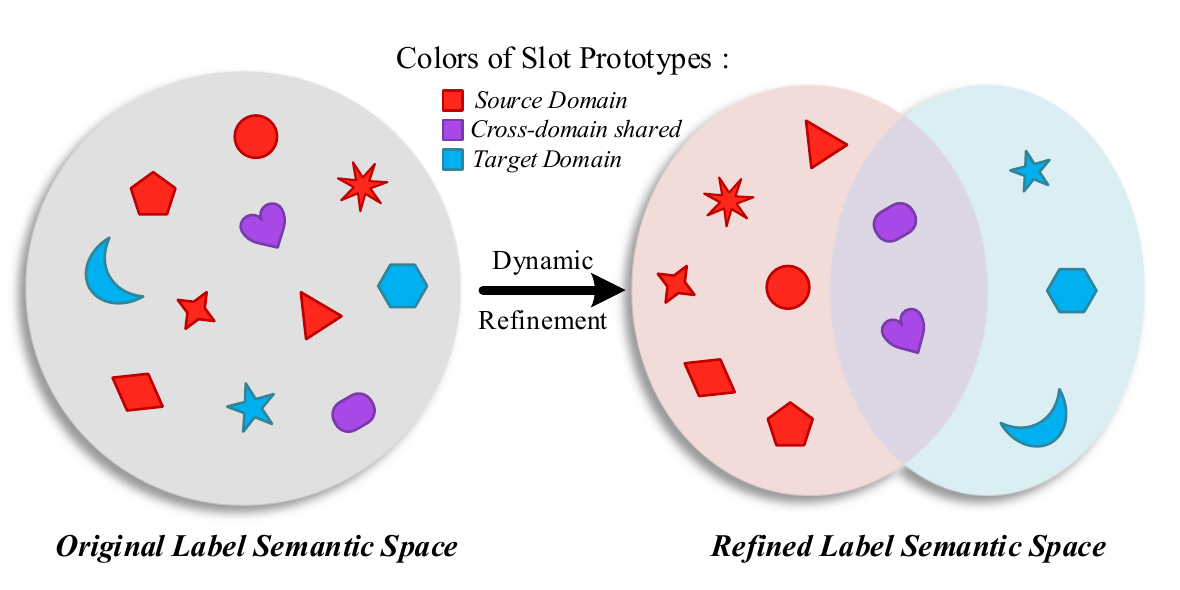}}
}
\vspace{-0.2cm} 
\caption{\label{fig:intro}(a) Performance of the models in \emph{GetWeather} domain; (b) a demonstration of slot prototypes refinement process;}
\vspace{-0.6cm} 
\end{figure}

However, we find that these methods have poor performance on unseen slot in the target domain,  as shown in Fig \ref{fig:intro}(a). In the cross-domain slot filling task, there are always seen slots and unseen slots in the target domain. The former exists in both the source domain and the target domain, while the latter only exists in the target domain. Although previous methods have achieved good overall performance on the cross-domain slot filling task, we find that their high performance mostly comes from the seen slots, while the performance on the unseen slots remains very low. We argue that these methods don't achieve domain adaption well. Actually, these methods lack explicit modeling of the association between the source and target domain. They directly utilize the slot name embedding as its slot prototype, whose distribution is often chaotic in semantic space due to lack of modeling for constraint relationships of slot prototypes in different domains. And due to the lack of data in the target domain, the model can't learn the mapping relationship between the slot value in the target domain and the slot prototype. Therefore, when making predictions,  the model can only correctly predict the seen slot type, and the prediction for unseen slots is almost random. An intuitive way to solve this problem is to refine the label semantic space and establish the constraint relationship between source and target domains, as shown in Fig \ref{fig:intro}(b).
\vspace{-0.05cm}

In this paper, we propose a novel method based on \textbf{P}rototypical \textbf{C}ontrastive learning and \textbf{L}abel \textbf{C}onfusion strategies \textbf{(PCLC)} to dynamically refine the constraint relationship between slot prototypes in the semantic space. First, we introduce prototypical contrastive learning \cite{medina2020selfsupervised,cao2020unsupervised,li2020contrastive,yue2021prototypical,li2021prototypical,zhao2021contrastive}, which makes the mapped slot value embeddings close to its corresponding slot prototype and away from other slot prototypes, to enhance the accuracy of mapping between feature space and semantic space. Second, we introduce a label obfuscation strategy to establish the dependency between the slots of the source domain and the target domain. In the training process of the source domain, we confuse the original one-hot label into the probability distribution of the source domain and the target domain by calculating the similarity between the slot prototypes in the source domain and the target domain.

Our contributions are three-fold. (1) We evaluate the performance of existing methods on cross-domain slot filling. It turns out these methods do not achieve domain adaptation effectively as the performance varies widely between unseen slots and seen slots. (2) We propose a novel method based on prototypical contrastive learning and label confusion to refine semantic space and enhance the domain adaptation. (3) Experiments demonstrate that the performance of our proposed method has improved significantly on unseen slots, and the overall performance outperforms the state-of-the-art models on both zero-shot and few-shot settings.
\vspace{-0.6cm}

\section{Method}
\vspace{-0.2cm} 
\begin{figure}[t]
\centering
\resizebox{.48\textwidth}{!}{\includegraphics{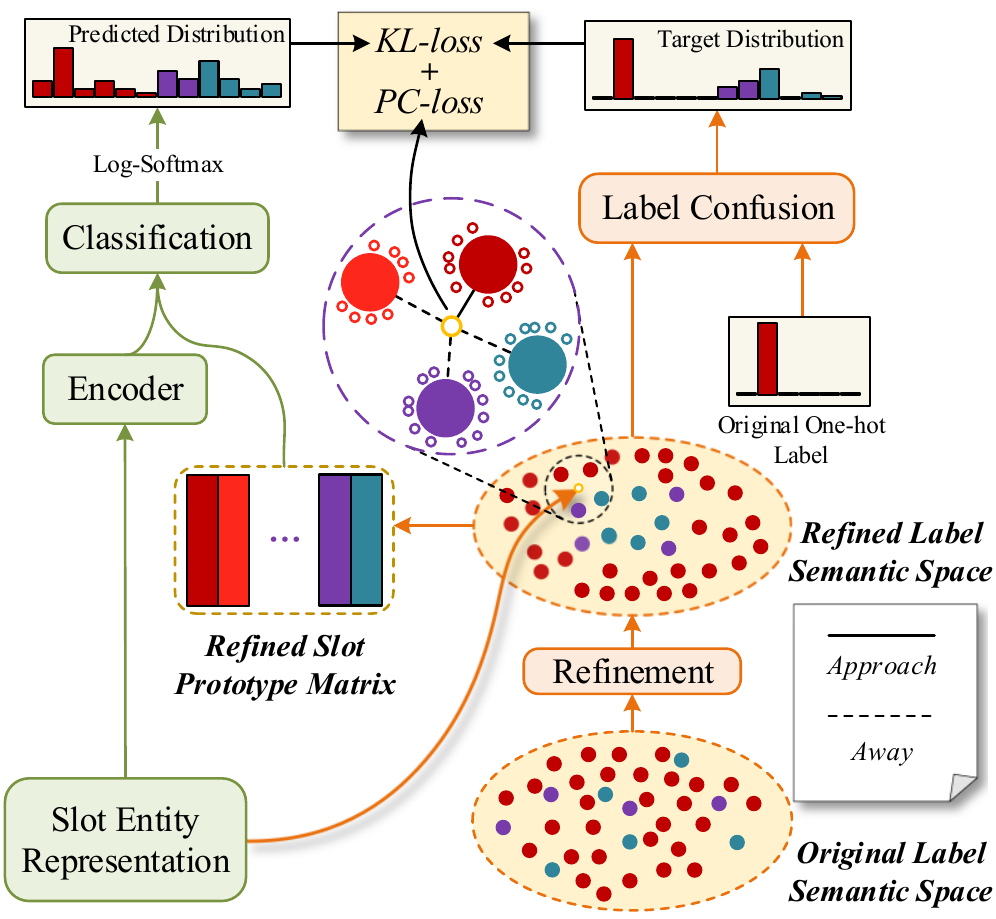}}
    \caption{Illustration of proposed method \bf{\emph{PCLC}}}
\label{fig:method}
% \vspace{-0.6cm}
\vspace{-0.3cm} 
\end{figure}

\subsection{Overall Architecture}
Consistent with Coach \cite{liu2020coach}, we adapt the two-stage framework as the backbone of our model. In the first stage, we utilize a BiLSTM-CRF structure \cite{lample-etal-2016-neural} (as a BIO-label sequence tagger) to encode the input utterance and identify the slot entities. 

In the second stage, our model encodes the slot entity and predicts the label for it by calculating the similarity with the slot prototypes in the label semantic space. 
Unlike the previous works \cite{liu2020coach,he-etal-2020-contrastive} which directly utilize the slot name embedding as slot prototypes, we introduce Prototypical Contrastive learning and Label Confusion strategies (PCLC) strategies to dynamically refine the constraint relationship between slot prototypes in the semantic space, as shown in Fig \ref{fig:method}. In the training procedure, we use an MLP layer to encode the original slot name embedding. So we can obtain a dynamically updated slot prototype matrix.
We will introduce these in the following subsections.
\vspace{-0.1cm}

\subsection{Prototypical Contrastive Learning}
There is a problem of utilizing slot name embedding as the slot prototypes that the distribution of slot name embedding is very chaotic and somewhat dense in semantic space. Therefore, when the slot values are mapped to the semantic space, they can hardly establish a correct relationship with the corresponding slot prototype. So, we introduce prototypical contrastive learning \cite{li2021prototypical} to enhance the precision of the mapping function from feature space to semantic space and reduce the density of slot prototype distribution in the label semantic space.

As shown in Fig \ref{fig:method}, given a slot value $r_k$ and the corresponding label $y_k$, we will map $r_k$ to the refined semantic space, and obtain the prototypical contrastive loss:
\begin{equation}
\setlength\abovedisplayskip{1pt}
    \mathcal{L}_{\mathrm{pc}}\left({r}_{k}, y_{k}\right)=-\log \frac{\exp \left({r}_{k} \cdot \boldsymbol{z}^{y_{k}} / \tau\right)}{\sum_{c=1}^{C} \exp \left({r}_{k} \cdot \boldsymbol{z}^{c} / \tau\right)}
\end{equation}
\vspace{-0.2cm}

, where $z^c$ is the representation of slot prototype with label $c$, $\tau$ is temperature factor and $C$ is the number of slot categories. By optimizing the above objective function $\mathcal{L}_{\mathrm{pc}}$, slot values can be close to corresponding slot prototype in semantic space and be away from other slot prototypes.
\subsection{Label Confusion}
Now that prototypical contrastive learning has separated the slot prototypes and established the relation between slot values and slot prototypes, we need to establish the dependency between the slots of the source domain and the target domain by label confusion. 

In the training process of the source domain, we confuse the original one-hot label into the probability distribution of the source domain and the target domain by calculating the similarity between the slot prototypes in the source domain and the target domain. Given a slot value $r$ and its corresponding label $y$, we first calculate the cosine similarities between $\boldsymbol{z}^{y}$ and slot prototypes $\boldsymbol{z}^{j}$ of all slot label $j \in \mathcal{T}$ in the target domain:
\vspace{-0.2cm}
\begin{gather}
    s^j ={\rm cosine\_similarity}(\boldsymbol{z}^{y}, \boldsymbol{z}^{j}) \\
    \mathcal{D}_\text{tgt} = \text{norm}( \{ s^j \}_{j \in \mathcal{T}} )
\end{gather}

\vspace{-0.1cm}
, where the $\text{norm}(\cdot)$ indicates the $\text{L}_1$ norm. 
We regard the $\mathcal{D}_\text{tgt}$ as the soft distribution over the slot labels in the target domain and then concatenate them with the one-hot distribution on the source domain with label confusion factor $\lambda$:
\vspace{-0.2cm}
\begin{gather}
    \mathcal{D}_\text{src} = \text{one\_hot}(y) \\
    \mathcal{D}_\text{smooth} = \text{concat}(\lambda \cdot \mathcal{D}_\text{src}, (1 - \lambda) \cdot \mathcal{D}_\text{tgt})
\end{gather}

\vspace{-0.1cm}
, where $\text{one\_hot}(y)$ indicates the one-hot distribution of the source slot label $y$ and $\text{concat}(\cdot)$ indicates the concatenation operation. Then we can obtain the KL-Divergence loss:
\vspace{-0.3cm}
\begin{gather}
    \mathcal{D}_{pre}= {\rm log\_softmax}(\mathcal{M}_{proto}\cdot r_k)\\
\mathcal{L}_{kl} = KL \text { -Divergence }(\mathcal{D}_{pre},\mathcal{D}_{smooth})
\end{gather}

\vspace{-0.2cm}
, where $\mathcal{M}_{proto}$ donate the representation matrix of the slot prototypes. The final loss function is $\mathcal{L}=\mathcal{L}_{pc}+\alpha \mathcal{L}_{kl}$, where
$\alpha$ is a hyperparameter.

\section{Experiments}

\begin{table*}[t]
\centering
\resizebox{\textwidth}{!}{
\begin{tabular}{c|cccc|ccc|cccc|ccc}  
\hline
Training Setting & \multicolumn{7}{c|}{Zero-shot} & \multicolumn{7}{c}{Few-shot on 50 samples} \\
\hline
Domain $\downarrow$ $\sim$ Model $\rightarrow$ & CT & RZT & Coach & CZSL-Adv & LC & PCL & PCLC & CT & RZT & Coach & CZSL-Adv & LC & PCL & PCLC \\
\hline
AddToPlaylist & 38.82& 42.77& 50.90& 53.89& 53.53& 55.60 & \textbf{59.24} & 68.69& 74.89& 74.68& 76.18& 81.55& 79.46& \textbf{81.59}\\
BookRestaurant & 27.54& 30.68& 34.01& 34.06& 34.89& 35.28& \textbf{41.36}& 54.22 & 54.49 & 74.82 & 76.28 & 76.48& 76.87& \textbf{77.54}\\
GetWeather & 46.45 & 50.28 & 50.47 & 52.24 & 49.16& 48.61& \textbf{54.21}& 63.23 & 58.87 & 79.64 & 83.28 & 83.83& 79.61& \textbf{83.85}\\
PlayMusic & 32.86 & 33.12 & 32.01 & 34.59 & 29.83& 33.02& \textbf{34.95}& 54.32 & 59.20 & 66.38 & \textbf{68.17} & 66.25& 67.48& 66.86\\
RateBook & 14.54 & 16.43 & 22.06 & \textbf{31.53} & 22.20& 23.25& 29.31& 76.45 & 76.87 & 84.62 & 87.22 & 87.14& \textbf{89.06}& 87.88\\
SearchCreativeWork & 39.79 & 44.45 & 46.65 & 50.61 & 48.73& \textbf{53.73}& 53.51& 66.38 & 67.81 & 64.56 & 66.49 & 69.63& 70.88& \textbf{71.06}\\
FindScreeningEvent & 13.83 & 12.25 & 25.63 & \textbf{30.05} & 20.99& 23.99& 27.17& 70.67 & 74.58 & \textbf{83.85} & 83.26 & 82.32& 83.46& 81.56\\
\hline
Average F1 & 30.55 & 32.85 & 37.39 & 40.99 & 37.05& 39.07& \textbf{42.82}& 64.85 & 66.67 & 75.51 & 77.27 & 78.17& 78.12& \textbf{78.62}\\
\hline
\end{tabular}
}
\vspace{-0.8pt}
\caption{Slot F1-scores on SNIPS for different target domains under zero-shot and few-shot learning settings. LC denotes adding label confusion strategy and PCL denotes adding prototypical contrastive loss.}
\label{tbl:control-results1}
\vspace{-0.4cm} 
\end{table*}

\subsection{Setup}

%Footnotes are inserted with the \verb|\footnote| command.\footnote{This is a footnote.}
\textbf{Dataset.}
We evaluate our method on SNIPS \cite{coucke2018snips}, a public spoken language understanding dataset which contains 7 domains and 39 slots. To simulate the cross-domain scenarios, we follow the setup of \cite{liu2020coach}, choosing one domain as the target domain and the left six domains as the source domains.

%To prove the effectiveness of our method, we evaluate our method on the Snips(),a public spoken language understanding dataset which contains 39 unique slots across seven domains with 2000 training examples approximately per domain. During the implementation process, we found some statistical mistakes in the previous work(), so we re-recorded the detailed statistics of Snips dataset in the appendix.  To test our framework, we follow the setup of (), choosing one domain as the target domain and the other six domains as the source domains.

\textbf{Baselines.}
We compare our approach with the following baselines:
\begin{itemize}
    \item \textbf{Concept Tagger (CT)} A method proposed by \cite{bapna2017zeroshot}, which utilizes the slot description to boost the performance on the unseen slots in the target domain.
    
    \item \textbf{Robust Zero-shot Tagger (RZT)} Based on CT, a method  proposed by \cite{shah-etal-2019-robust}, which adds additional example values of  slots to improve the robustness of the model.
    
    \item \textbf{Coarse-to-fine Approach (Coach)} A two-stage framework proposed by \cite{liu2020coach}: the first stage performs coarse-grained BIO labeling task, and the second stage utilizes slots descriptions to perform fine-grained slot type classification task. And it has a stronger variant\footnote{For brevity, all the \emph{Coach} we mentioned in subsequent experiments refer to this stronger variant.}, which applies template regularization to improve performance.
    
    \item \textbf{Contrastive Zero-Shot Learning with Adversarial Attack (CZSL-Adv)} A method proposed by \cite{he-etal-2020-contrastive} based on Coach, which utilizes contrastive learning and adversarial attacks to improve the performance and robustness of the framework.
\end{itemize}
\vspace{-0.4cm} 
\textbf{Implementation Details}
We follow the setup of \cite{liu2020coach}, selecting one domain as the target domain at a time, and use  500 samples in this domain as a validation set, the rest as a test set. Samples from the remaining six domains are used for training. We use a two-layer BiLSTM with a hidden layer size of 200 as encoder, and concatenate the word-level and character-level embedding as input. The dropout rate of BiLSTM is set to 0.3 and the learning rate of the Adam optimizer is set to 0.0005. We set the batch size to 64 and use the early stop of patience 15 to ensure the stability of the model. For all the experiments, we train and test our model on the 2080Ti GPU. It takes an average of 2.5 hours to run with 30 epochs under the zero-shot setting and 5 hours to run with 60 epochs under the few-shot setting.

% Our model achieves the best performance when label confusion factor $\lambda$ is set to 0.6 and $\alpha$ set to 0.5.

% We use a two-layer BiLSTM with a hidden layer size of 200 as the encoder, and concatenate word-level and character-level embedding as the final embeddings of the input token. The dropout rate of BiLSTM is set to 0.3 and the learning rate of the Adam optimizer is set to 0.0005. We set batch size to 64 and use the early stop of patience 15 to ensure the stability of the model.

% For all the experiments, we train and test our model on the 2080Ti GPU. The total parameters of our models are about 15M under both zero-shot and few-shot settings. It takes an average of 2.5 hours to run with 30 epochs under zero-shot setting and 5 hours to run with 60 epochs under few-shot setting.
\vspace{-0.3cm}

\subsection{Main Results}

As illustrated in Table \ref{tbl:control-results1}, our method PCLC outperforms the SOTA model by 1.83\% on the average F1-score under zero-shot setting, and 1.35\% under few-shot setting. Besides, compared to Coach, the model we directly modify, our method achieves superior performance by 4.7\% under zero-shot setting and 3.1\% under few-shot setting. The significant performance improvement proves that the combined use of the two strategies we proposed can help establish a better mapping relation between slots values and slot prototypes in label semantic space.

\subsection{Results on Seen slots and Unseen Slots}
In this section, we re-explore all these methods' transferability by measuring their performance directly on unseen slots and seen slots respectively\footnote{\newcite{liu2020coach} split SNIPS into \emph{seen} and \emph{unseen} part according to whether an utterance includes the unseen slots, which may introduce some bias to the actual performance when testing model on unseen slots.}. The experimental results are shown in Table \ref{tbl2:control-results}. We can observe that the previous methods perform very limited on the unseen slot under the zero-shot setting, while PCLC has achieved a very significant improvement on the unseen slot under both settings. For huge improvement on unseen slots, we hypothesize that with our proposed two strategies our model does achieve a more effective knowledge transferring to the target domain instead of overfitting the seen slot. It is necessary to mention that there still exists huge gap of model performance on unseen and seen slot, which is worth further study. More details about all domains can be found in Table \ref{tbl3:control-results}.
\vspace{-0.2cm}

\begin{table}[t]
\centering
\resizebox{.40\textwidth}{!}{
\begin{tabular}{c|cc|cc}  
\hline
\multirow{2}*{settings} & \multicolumn{2}{c|}{0 samples} & \multicolumn{2}{c}{50 samples} \\
\cline{2-5}
& unseen & seen & unseen & seen\\
\hline
CT & 3.38 & 37.23 & 52.65& 65.66\\
RZT & 2.19 & 40.99 & 50.28& 61.63\\
Coach & 9.31 & 46.22 & 68.59& 74.55\\
\hline
LC & 9.15 & 48.65 & 75.09& \textbf{79.60}\\
PCL & 10.71 & 49.08 & 74.78& 79.44\\
PCLC & \textbf{17.38} & \textbf{51.68} &\textbf{76.64} &78.99 \\
\hline
\end{tabular}
}
\vspace{-0.2cm} 
\caption{Average F1-scores for seen and unseen slots across all target domains \footnotemark[3]}
\label{tbl2:control-results}
\vspace{-0.6cm} 
\end{table}

\subsection{Ablation Analysis}
\vspace{-0.1cm}

\begin{table*}[t]
\centering
\resizebox{\textwidth}{!}{%
\begin{tabular}{c|cc|cc|cc|cc|cc|cc|cc|cc}  
\hline
samples & \multicolumn{8}{c|}{Zero-shot} & \multicolumn{8}{c}{Few-shot on 50 samples} \\
\hline
\multirow{2}*{model} & \multicolumn{2}{c|}{CT} & \multicolumn{2}{c|}{RZT} & \multicolumn{2}{c|}{Coach} &
\multicolumn{2}{c|}{PCLC} & \multicolumn{2}{c|}{CT} & \multicolumn{2}{c|}{RZT} & \multicolumn{2}{c|}{Coach} &
\multicolumn{2}{c}{PCLC} \\
\cline{2-17} & unseen& seen & unseen& seen & unseen& seen & unseen& seen & unseen& seen & unseen& seen & unseen& seen & unseen& seen \\
\hline
AddToPlaylist & 5.18 & 47.15 & 3.48 & 57.50 & \textbf{7.94} & 64.65 & 2.57 & \textbf{73.32} & 60.12 & 73.87 & 54.71 & 68.72 & 66.11 & 76.26 & \textbf{75.16} & \textbf{84.60}\\
BookRestaurant & 1.87 & 51.43 & 7.14 & 43.84 & 2.89 & 55.88 & \textbf{16.56} & \textbf{62.81} & 46.25 & 56.83 & 47.40 & 52.35 & 64.94 & 79.15 & \textbf{73.30} & \textbf{83.16}\\
GetWeather & 2.11 & 39.54 & 2.34 & 62.84 & 2.20 & 63.97 & \textbf{14.20} & \textbf{65.84} & 31.91 & 71.61 & 29.46 & 68.83 & 59.71 & 84.15 & \textbf{75.12} & \textbf{88.02}\\
PlayMusic & 0.13 & 46.48 & 0 & \textbf{47.45} & 8.78 & 31.69 & \textbf{17.53} & 45.17 & 35.36 & 62.27 & 35.21 & 61.80 & 54.24 & 68.02 & \textbf{62.79} & \textbf{69.56}\\
RateBook & 0.13 & 25.10 & 0.13 & 25.41 & 18.81 & \textbf{40.42} & \textbf{25.70} & 34.70 & 81.39 & 58.69 & 80.65 & 55.64 & 89.11 & 70.38 & \textbf{93.15} & \textbf{72.98}\\
SearchCreativeWork & - & 39.59 & - & 39.27 & - & 44.35 & - & \textbf{53.51} & - & 64.69 & - & 39.27 & - & 58.49 & - & \textbf{71.05}\\
SearchScreeningEvent & 10.84 & 11.32 & 0 & 10.61 & 15.21 & 27.27 & \textbf{22.71} & \textbf{29.66} & 60.82 & 71.67 & 54.21 & 56.43 & 77.41 & \textbf{85.36} & \textbf{80.29} & 83.52\\
\hline
Avg & 3.38 & 37.23 & 2.19 & 40.99 & 9.31 & 46.22 & \textbf{17.38} & \textbf{51.68} & 52.65 & 65.66 & 50.28 & 61.63 & 68.59 & 74.55 & \textbf{76.64} & \textbf{78.99}\\
\hline
\end{tabular}}
\caption{Detailed F1-scores for seen and unseen slots across all target domains. The bold part is the highest F1-score obtained by our model and baselines on the unseen and seen slots, under the zero-shot and the few-shot settings respectively.}
\label{tbl3:control-results}
\vspace{-0.4cm} 
\end{table*}

We compare the effect for the main performance of PCL and LC strategy in Table \ref{tbl:control-results1}. We find that the LC strategy would cause a slight decrease in performance under the zero-shot setting while PCL can provide some performance boost, comparing to coach. Interestingly, both LC and PCL can achieve a significant improvement close to PCLC under the few-shot setting. Besides, we conduct the ablation experiment on unseen slots, as shown in Table \ref{tbl2:control-results}. We find that LC and PCL don't work very well individually on unseen slots under zero-shot setting. The experimental results show that both PCL and LC would bring significant improvement under the few-shot setting, but they need to be combined together for better performance under the zero-shot setting.
\vspace{-0.2cm}

\subsection{Visualization Analysis}

\begin{figure}[t]
\centering
\resizebox{.48\textwidth}{!}{\includegraphics[scale=0.1754]{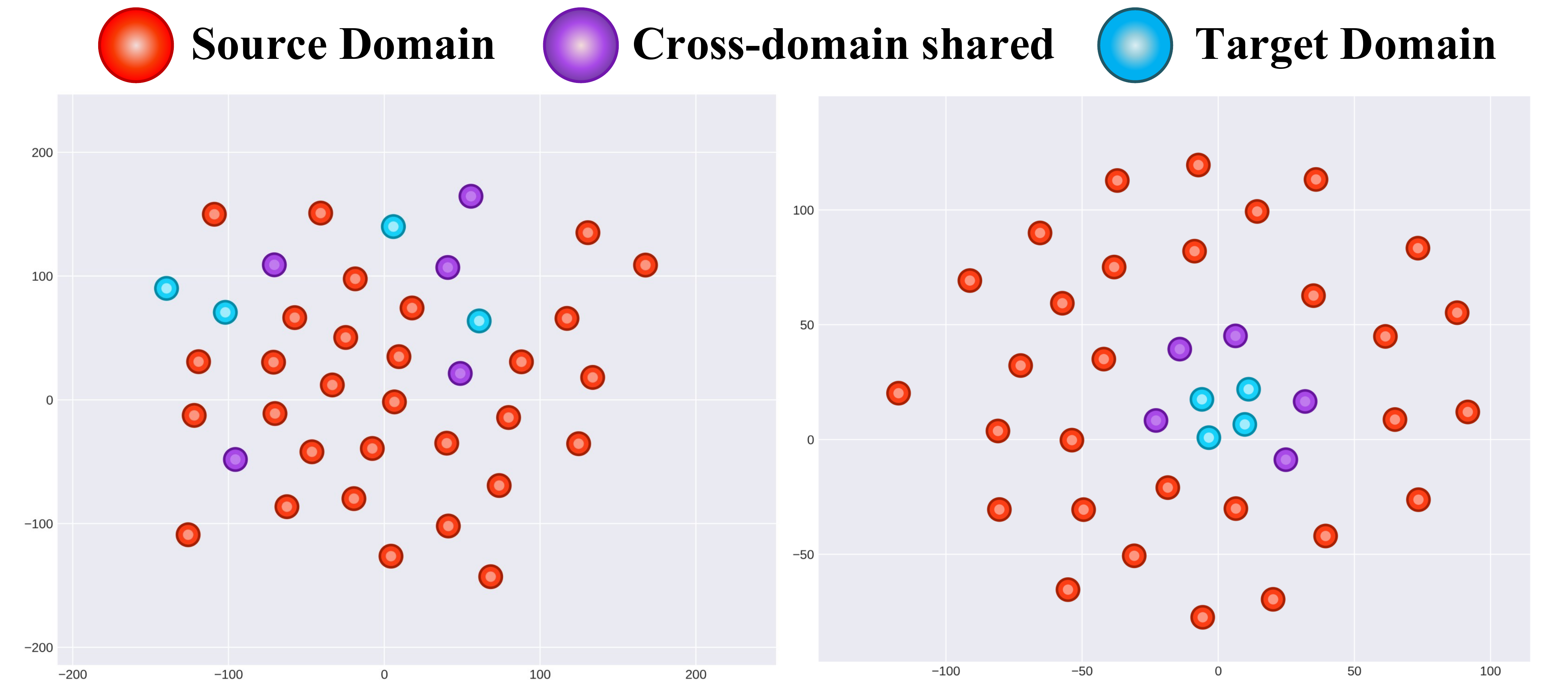}
}
\caption{t-SNE projection of slot prototypes. The left represents the original distribution and the right represents refined distribution by our method. The target domain is \emph{Getweather} while others are source domains.}
\label{Visualization}
\vspace{-0.5cm} 
\end{figure}

To explore the effectiveness on the refinement of the slot prototype embedding in semantic space, we do the t-SNE visualization for the slot prototype representation after refinement, as shown in Fig \ref{Visualization}. It is observed that after refinement, the distribution of slot prototype in semantic space changes from a chaotic distribution to a separated distribution between the source domain and the target domain. This observation indicates our thoughts: prototypical contrastive learning stimulates different slots to be separated from each other, while label confusion is able to establish the constraint relations between the source domain and the target domain. The separation between the source domain and the target domain helps to build the mapping between slot value and slot prototype, and thus improve the accuracy on unseen slots.
% \vspace{-0.3cm}

\vspace{-0.2cm} 
\subsection{Hyperparameter Analysis}

\begin{figure}[h]
\centering
\resizebox{.48\textwidth}{!}{\includegraphics{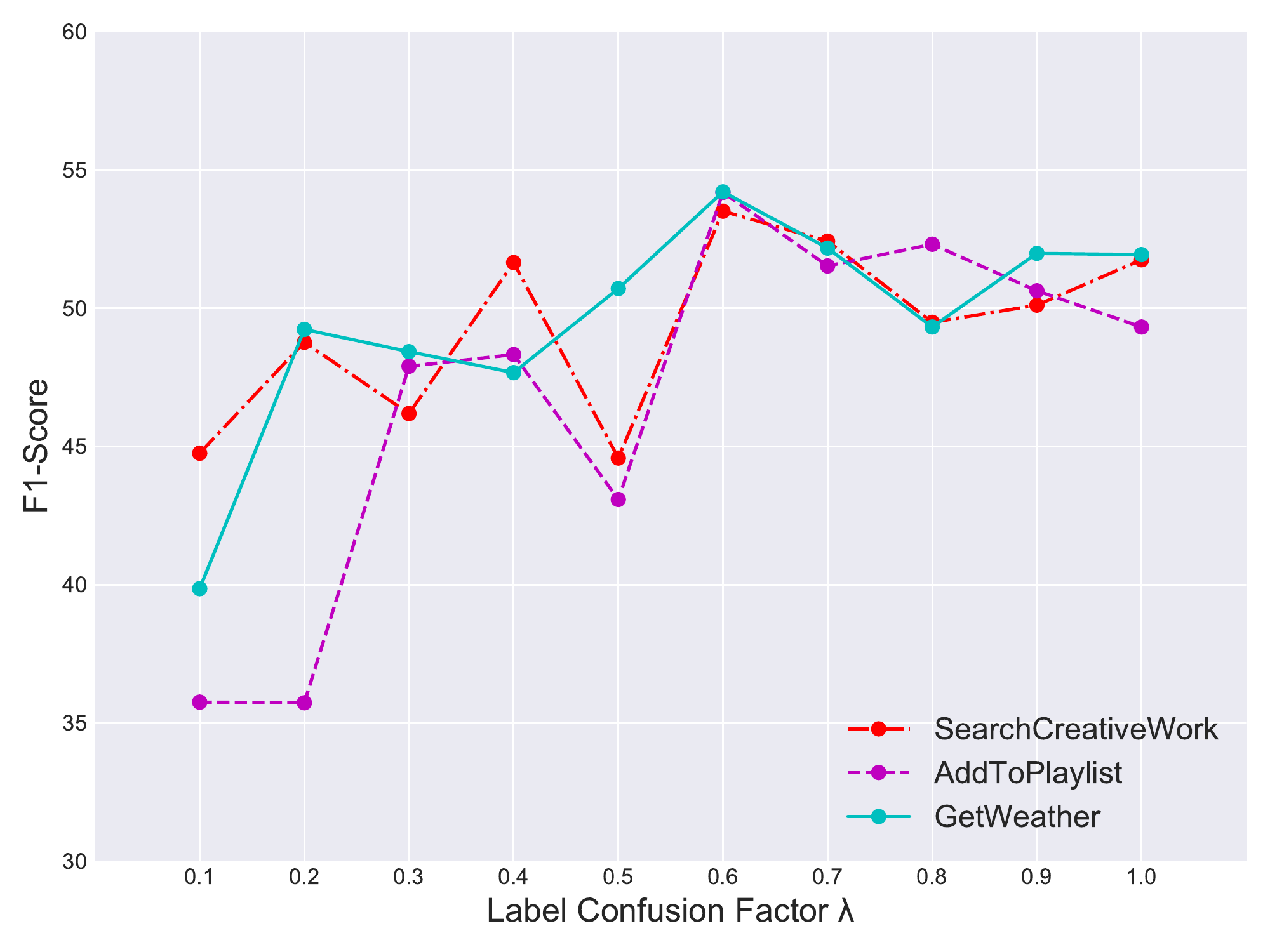}}
\caption{\label{fig:smooth}F1-scores of PCLC with different label confusion factor $\lambda$.}
\vspace{-0.5cm} 
\end{figure}

The hyperparameter $\lambda$ is used to control the degree of label confusion, which is set too small will make the prediction too random, while set too large the original effect will not be achieved. As shown in Figure \ref{fig:smooth}, the model has the best performance when we set $\lambda$ to 0.6
\vspace{-0.2cm} 
\section{Conclusion \& Discussion}
\label{sec:bibtex}
In this paper, we propose a novel method based on Prototypical Contrastive learning and Label Confusion strategy (PCLC) for cross-domain slot filling. Our main contribution was to improve the domain adaptability of the model. The proposed method conducts a refinement process for label semantic space to re-establish the constraint relationship between different slots. Experiments show the effectiveness of our method, especially for recognizing the unseen slots.

As there is still a huge gap in model performance between unseen and seen slots, in future work, we will focus on improving the performance on unseen slots while maintaining the performance on seen slots. Representation disentanglement \cite{wang-etal-2021-dynamically} can be used to disentangle domain-specific and domain-shared knowledge in the source domain, then we can preserve domain-shared knowledge and focus on establishing the relation of domain-specific knowledge between the source and the target domain.

\vspace{-0.2cm} 
\section{Acknowledgements}
\label{sec:bibtex}
We thank Jinzheng Zhao, Yuanyuan Qi and all anonymous reviewers for their helpful comments and suggestions. This work was partially supported by National Key R\&D Program of China No. 2019YFF0303300 and Subject II  No. 2019YFF0303302, DOCOMO Beijing Communications Laboratories Co., Ltd, MoE-CMCC "Artifical Intelligence" Project No. MCM20190701.

% Entries for the entire Anthology, followed by custom entries
\bibliography{custom}
\bibliographystyle{acl_natbib}

\end{document}